\UseRawInputEncoding
\documentclass{article}

\usepackage[preprint]{neurips_2025}

\usepackage[utf8]{inputenc} % allow utf-8 input
\usepackage[T1]{fontenc}    % use 8-bit T1 fonts
\usepackage{hyperref}       % hyperlinks
\usepackage{url}            % simple URL typesetting
\usepackage{booktabs}       % professional-quality tables
\usepackage{amsfonts}       % blackboard math symbols
\usepackage{nicefrac}       % compact symbols for 1/2, etc.
\usepackage{microtype}      % microtypography
\usepackage{xcolor}         % 
\usepackage{tcolorbox}
\usepackage{tikz}
\usetikzlibrary{positioning}
\usetikzlibrary{calc}
\usepackage{amsthm}
\usepackage{amsmath}
\usepackage{algorithm}
\usepackage[noend]{algorithmic}
\usepackage{multirow}
\usepackage{todonotes}
\usepackage{wrapfig}

\newcommand{\torcs}{\textsc{Torcs}}
\newcommand{\karel}{\textsc{Karel}}
\newcommand{\parking}{\textsc{Parking}}

\newcommand{\psm}{\textsc{Psm}}
\newcommand{\ndps}{\textsc{Ndps}}
\newcommand{\leaps}{\textsc{Leaps}}
\newcommand{\propel}{\textsc{Propel}}
\newcommand{\daggeralg}{\textsc{Dagger}}

\usepackage{listings}
\usepackage{xcolor}

\newtcolorbox{mycommentbox}{
    colback=gray!5!white, colframe=gray!75!black,
    boxrule=0.5pt, arc=0mm, left=2pt, right=2pt, top=-1pt, bottom=-1pt,
    fontupper=\small
}

\title{Common Benchmarks Undervalue the Generalization Power of Programmatic Policies}

\author{%
  Amirhossein Rajabpour\textsuperscript{1} \quad
  Kiarash Aghakasiri\textsuperscript{1} \quad
  Sandra Zilles\textsuperscript{2} \quad
  Levi H.\,S.\,Lelis\textsuperscript{1}\\
  \textsuperscript{1}Amii, Department of Computing Science, University of Alberta \\ \textsuperscript{2}Department of Computer Science, University of Regina 
}

\begin{document}

\maketitle

\begin{abstract}
Algorithms for learning programmatic representations for sequential decision-making problems are often evaluated on out-of-distribution (OOD) problems, with the common conclusion that programmatic policies generalize better than neural policies on OOD problems. In this position paper, we argue that commonly used benchmarks undervalue the generalization capabilities of programmatic representations. We analyze the experiments of four papers from the literature and show that neural policies, which were shown not to generalize, can generalize as effectively as programmatic policies on OOD problems. This is achieved with simple changes in the neural policies training pipeline. Namely, we show that simpler neural architectures with the same type of sparse observation used with programmatic policies can help attain OOD generalization. Another modification we have shown to be effective is the use of reward functions that allow for safer policies (e.g., agents that drive slowly can generalize better). Also, we argue for creating benchmark problems highlighting concepts needed for OOD generalization that may challenge neural policies but align with programmatic representations, such as tasks requiring algorithmic constructs like stacks.
\end{abstract}

\section{Introduction}
Deep reinforcement learning (RL) has led to remarkable successes in domains ranging from games to robotics, largely by representing policies as highly parametrized neural networks and optimizing them end‐to‐end \citep{lillicrap2019continuouscontroldeepreinforcement,schulman2017proximal}.  However, neural policies often struggle to generalize outside the distribution of their training environments, exhibiting brittle behavior when confronted with out‐of‐distribution (OOD) scenarios.  
In contrast, a growing literature on \emph{programmatic} policies, where decision–making rules are expressed in a domain‐specific language, claims superior OOD generalization \citep{pirl,verma2019imitation,leaps,InalaBTS20}.

We argue that commonly used benchmarks undervalue the generalization power of programmatic representations. 
Previous work on programmatic policies has observed a substantial gap in terms of OOD generalization between programmatic and neural representations. We revisit these OOD generalization claims and show that, in some cases, the apparent gap between programmatic and neural representations arises not from an inherent limitation of neural representations but from variables we failed to control in evaluating OOD generalization with neural models. 
Namely, the input observation of the neural agent must be as sparse as the observation the programmatic agent considers. Sparse observations automatically remove distractions that can improve OOD generalization, especially when used with simpler models that are easier to train, such as fully connected networks. Moreover, neural policies tend to be more sensitive to the reward function because they tend to optimize it better than programmatic ones. As a result, a policy that is ``too specialized'' in one setting might perform poorly in OOD problems. As we show in our experiments, simple changes to the reward function can dramatically enhance OOD generalization. 

We demonstrate how some of these ideas improve the OOD generalization on benchmark problems commonly used in the literature, including a car racing environment (\torcs)~\citep{pirl,verma2019imitation}, grid‐world planning problems (\karel)~\citep{leaps}, and continuous control with repetitive behavior (\parking)~\citep{InalaBTS20}. Given our observation that neural policies can generalize to OOD problems in these benchmarks, we suggest creating problems that showcase the OOD generalization of programmatic representations by requiring learning structures that neural networks fail to master, such as stacks~\citep{joulin2015inferring}. As an illustrative example, we suggest a problem that requires the agent to use memory through a stack or a queue to solve. 

Focusing on benchmark problems that require features beyond the reach of neural models will help us better understand where programmatic representations are most needed. This understanding can help us develop novel representations that combine the flexibility of highly parameterized models with the desired properties of symbolic programs, such as sparsity and the usage of complex data structures.  

The code used to run our experiments is publicly available \href{https://github.com/lelis-research/Common-Benchmarks-Undervalue-the-Generalization-Power-of-Programmatic-Policies}{online}.

\section{Problem Definition}

We consider sequential decision-making problems as Markov decision processes (MDPs) $\mathcal{M} = (S, A, p, r, \mu, \gamma)$. Here, $S$ and $A$ are the sets of states and actions. The function $p : S \times A \rightarrow S$ is the transition model, which returns the state $s_{t+1}$ reached once the agent takes action $a_t$ in state $s_t$ at time step $t$. The agent observes a reward value of $R_{t+1}=r(s_t,a_t)$ when transitioning to $s_{t+1}$; such values are given by the reward function $r : S \times A \rightarrow \mathbb{R}$. The MDP's initial states are determined by the distribution $\mu$, with states sampled from $\mu$ denoted as $s_0$. Finally, $\gamma\in [0, 1]$ is the discount factor. A policy $\pi : S \times A \rightarrow [0, 1]$ receives a state $s$ and action $a$ and returns the probability of taking $a$ at $s$. Given a class of policies $\Pi$, the goal is to find a policy $\pi$ within $\Pi$ that maximizes the return:

\begin{equation}
\operatorname*{arg\,max}_{\pi \in \Pi} \mathbb{E}_{\pi,p,\mu}[\sum_{k=0}^\infty \gamma^k R_{k+1}]
\label{eq:rl_problem}
\end{equation}
The class $\Pi$ determines the biases of the policies we consider. For example, $\Pi$ could be an architecture of a neural network, and the policies $\pi$ within this class are the different weights we can assign to the connections of the neural network. We consider classes $\Pi$ determined by a domain-specific language, so programs written in the language form $\Pi$. A language is defined with a context-free grammar $(\mathcal{N}, \mathcal{T}, \mathcal{R}, \mathcal{I})$, where $\mathcal{N}$, $\mathcal{T}$, $\mathcal{R}$, $\mathcal{I}$ are the sets of non-terminals, terminals, the production rules, and the grammar's initial symbol, respectively. Figure~\ref{fig:torcs} (a) shows an example of a context-free grammar encoding a language for \torcs\ policies. The grammar's initial symbol $\mathcal{I}$ is $E$. It accepts strings such as the one shown in Figure~\ref{fig:torcs} (b), which is obtained through a sequence of production rules applied to the initial symbol: $E \rightarrow \mathbf{if}~B~\mathbf{then}~E~\mathbf{else}~E \rightarrow \mathbf{if}~B~\mathbf{and}~B~\mathbf{then}~E~\mathbf{else}~E \rightarrow \cdots$. 

We empirically compare solutions to Equation~\ref{eq:rl_problem} when the class $\Pi$ is defined with pre-defined neural network architectures and domain-specific languages. We call the former neural and the latter programmatic policies. We consider the following problem domains in our experiments: \torcs~\citep{pirl,verma2019imitation}, \karel~\citep{leaps}, and \parking~\citep{InalaBTS20}. 

\section{Background: Searching for Programmatic Policies}

This section describes the algorithms used to synthesize programmatic policies for solving \torcs\ (Section~\ref{sec:ndps}), \karel\ (Section~\ref{sec:leaps}), and \parking\ (Section~\ref{sec:psm}). We aim to provide enough information so the reader understands our results in Section~\ref{sec:experiments}. We do not intend to detail the original algorithms. For full method descriptions, see the cited papers in each subsection. 

\subsection{Neurally Directed Program Search (\ndps)}
\label{sec:ndps}

\cite{pirl} introduced Neurally Directed Program Search (\ndps), a method that uses imitation learning through the \daggeralg\ algorithm~\citep{ross2011reduction} to learn programmatic policies. Figure~\ref{fig:torcs} (a) shows the domain-specific language \cite{pirl} considered in their experiments on the \torcs\ benchmark. The $\mathbf{peek}$ function reads the value of a sensor. For example, $\mathbf{peek}(h_\mathtt{RPM}, -1)$ reads the latest value (denoted by the parameter $-1$) of the rotation-per-minute sensor ($h_\mathtt{RPM}$); $\mathbf{peek}(h_\mathtt{RPM}, -2)$ would read the second latest value of the sensor. The $\mathbf{fold}(+, \epsilon- h_i)$ operation adds the difference $\epsilon- h_i$ for a fixed number of steps of the past readings of sensor $h_i$. 

The non-terminal symbols $P$, $I$, and $D$ in Figure~\ref{fig:torcs} (a) form the operations needed to learn PID controllers, with programs that switch between different PID controllers, as shown in Figure~\ref{fig:torcs} (b).

\begin{figure}[t]
    % \centering
    \begin{tabular}{ll}
    \textbf{(a) Domain-Specific Language} & \textbf{(b) Example Policy} \\
    \scalebox{0.9}{
    \begin{minipage}{0.45\textwidth}
        \begin{eqnarray*}
        P & ::= & \mathbf{peek}((\epsilon - h_i), -1) \\
        I & ::= & \mathbf{fold}(+, \epsilon - h_i) \\
        D & ::= & \mathbf{peek}(h_i, -2)- \mathbf{peek}(h_i, -1) \\
        C &::=& c_1 * P + c_2 * I + c_3 * D \\
        B & ::=  & c_0 + c_1 * \mathbf{peek}(h_1, -1) + \dots  \\
        & & \dots + c_k * \mathbf{peek}(h_m, -1) > 0 \mid \\
        & & \qquad B ~\mathbf{or}~ B \mid B ~\mathbf{and}~ B \\
        E  & ::= & C \mid \mathbf{if}~B~\mathbf{then}~E~\mathbf{else}~E. 
        \end{eqnarray*}
    \end{minipage}} 
    &
    \scalebox{0.85}{
    \begin{minipage}{0.45\textwidth}
        \begin{equation*}
        \begin{array}{l}
        \mathbf{if}~(0.001 - \mathbf{peek}(h_\mathtt{TrackPOS}, -1) > 0 ) \\ 
        \quad \mathbf{and}~(0.001 + \mathbf{peek}(h_\mathtt{TrackPOS}, -1) > 0) \\
        \quad \quad \mathbf{then}~ 3.97 * \mathbf{peek}((0.44 - h_\mathtt{RPM}), -1) \\
        \quad \quad \quad \quad +  0.01 * \mathbf{fold}(+, (0.44 - h_\mathtt{RPM})) \\ 
        \quad \quad \quad \quad + 48.79 * ( \mathbf{peek}(h_\mathtt{RPM}, -2)- \mathbf{peek}(h_\mathtt{RPM}, -1)) \\
        \quad \quad \mathbf{else}~ \;\, 3.97 * \mathbf{peek}((0.40 - h_\mathtt{RPM}), -1) \\ 
        \quad \quad \quad \quad + 0.01 * \mathbf{fold}(+, (0.40 - h_\mathtt{RPM})) \\ 
        \quad \quad \quad \quad +  48.79* ( \mathbf{peek}(h_\mathtt{RPM}, -2)- \mathbf{peek}(h_\mathtt{RPM}, -1))
        \end{array}
        \end{equation*}
    \end{minipage}}
    \\
    \end{tabular}
    \caption{(a) Context-free grammar specifying a domain-specific language for \torcs, a racing car domain~\citep{pirl}. The initial symbol of the language is $E$, $\epsilon$ is a pre-defined constant, and $\{h_i\}_{i=1}^m$ is a set of $m$ sensors from which the agent can read. The grammar allows programs that switch between different PID controllers. (b) Example of a policy written in the language.}
    \label{fig:torcs}
\end{figure}

\ndps\ uses a neural policy as an oracle to guide the \ndps's synthesis. Given a set of state-action pairs $H$, where the actions are given by the neural oracle, \ndps\ evaluates a program $\rho$ by computing the action agreement of $\rho$ with the actions in $H$. \ndps\ runs a brute force search algorithm~\citep{AlbarghouthiGK13,Udupa:2013}, to generate a set of candidate programs $C$. Then, it learns the parameters of the programs ($c_1$, $c_2$, and $c_3$ in Figure~\ref{fig:torcs}) with Bayesian optimization~\citep{snoek2012practical} such that the programs mimic $H$. Once \ndps\ determines the parameters of programs $C$, it selects the candidate $c$ in $C$ that maximizes the agent's return; $c$ is the starting point of a local search that optimizes a mixture of the action agreement function and the agent's return. 

\cite{verma2019imitation} introduced Imitation-Projected Programmatic Reinforcement Learning (\propel), an algorithm that also synthesizes a program for solving control problems. \propel\ is similar to \ndps\ in that it relies on a neural policy to guide its search through the space of programs. The difference between \propel\ and \ndps\ is that the neural policy of the former is trained so that it does not become ``too different'' from what the programmatic learner can express---the inability to represent the teacher's policy is known as the representation gap in the literature~\citep{qiu2021programmatic}. The programmatic policies of both \ndps\ and \propel\ are called for every state the agent encounters.

\subsection{Learning Embeddings for Latent Program Synthesis (\leaps)}
\label{sec:leaps}

\cite{leaps} introduced Learning Embeddings for Latent Program Synthesis (\leaps), a system that learns a latent representation of the space of programs a language induces. When given an MDP $\mathcal{M}$, \leaps\ searches in the learned latent space for a vector decoded into a program encoding a policy that maximizes the agent's return at $\mathcal{M}$. \leaps's premise is that searching in the learned latent space is easier than searching in the space of programs, as \ndps\ and \propel\ do. 

Figure~\ref{fig:karel_example} (a) shows the context-free grammar specifying the language used to encode policies for \karel. The language accepts programs with conditionals and loops. It also includes a set of perception functions, such as \texttt{frontIsClear}, which verifies whether the cell in front of the agent is clear. Further included are action instructions such as \texttt{move} and \texttt{turnLeft}. The set of perception functions is important because it defines what the agent can observe. As we show in Section~\ref{sec:experiments_karel}, having access to less information allows the agent to generalize to OOD problems. Figure~\ref{fig:karel_example} (b) shows an example of a \karel\ program. Here, the agent will perform two actions, \texttt{pickMarker} and \texttt{move}, if a marker is present in its current location; otherwise it will not perform any action. 

To learn its latent space, \leaps\ generates a data set of programs $P$ by sampling a probabilistic version of the context-free grammar defining the domain-specific language. That is, each production of a non-terminal can be selected with a given probability. A program can be sampled from this probabilistic grammar by starting at the initial symbol and randomly applying production rules until we obtain a program with only terminal symbols. This set of programs is used to train a Variational Auto-Encoder (VAE)~\citep{kingma2014autoencoding}, with its usual reconstruction loss. However, in addition to learn spaces that are more friendly to search algorithms, \leaps\ uses two additional losses that attempt to capture the semantics of the programs. These two losses incentivize latent vectors that decode into programs with similar agent behavior to be near each other in the latent space. The intuition is that this behavior locality can render optimization landscapes easier to search. 

\begin{figure}[t]
\begin{tabular}{ll}
    \textbf{(a) Domain-Specific Language} & \textbf{(b) Example Policy} \\
    \scalebox{0.9}{
    \begin{minipage}{0.65\textwidth}
    \[
    \begin{array}{l}
        \rho :=~\mathbf{def}~\texttt{run m(} s \texttt{ m)} \\
        s :=~ \mathbf{while}~\texttt{c(}b\texttt{ c) w( }s\texttt{ w) } | ~\mathbf{if}~\texttt{ c(}b\texttt{ c) i(}s\texttt{ i) } | \\
        \quad \quad \,\,\, \mathbf{ifelse}~\texttt{c(}b\texttt{ c) i(}s\texttt{ i)}~\mathbf{else}~\texttt{e(}s\texttt{ e) } |  \\ 
        \quad \quad \,\,\, \mathbf{repeat}~\texttt{ R=}n \texttt{ r(}s\texttt{ r) } | ~s;s\texttt{ }|\texttt{ }a \\
        b :=~ h\texttt{ }|~ \mathbf{not}~\texttt{(}h\texttt{)} \\
        n :=~ 0, 1, \cdots, 19 \\
        h :=~ \texttt{frontIsClear }|\texttt{ leftIsClear }|\texttt{ rightIsClear }| \\
        \quad \quad \,\,\,\texttt{markersPresent }|\texttt{ noMarkersPresent } \\
        a :=~ \texttt{move }|\texttt{ turnLeft }|\texttt{ turnRight }| \\
        \quad \quad \,\,\, \texttt{putMarker }|\texttt{ pickMarker }
    \end{array}
    \]
    \end{minipage}
    }
    &
    \scalebox{0.85}{
    \begin{minipage}{0.45\textwidth}
        \begin{tabular}[b]{l}
            $\mathbf{def}$ \texttt{run m(}\\
            \hspace*{0.5cm}$\mathbf{if}$ \texttt{c( markersPresent c) i(}\\
            \hspace*{0.5cm}\hspace*{0.5cm}\texttt{pickMarker move}\\
            \hspace*{0.5cm}\texttt{i)}\\
            \texttt{m)}
            \end{tabular}
    \end{minipage}
    }
    \end{tabular}
    \caption{(a) Context-free grammar specifying a domain-specific language for \karel. The programs written in this language accept conditional statements and loops. There is a set of perception functions ($h$) and functions that return actions ($a$). (b) Example of a policy for a \karel\ task.}
    \label{fig:karel_example}
\end{figure}

Once the latent space is trained, it is used to solve MDPs. Given an MDP, \leaps\ uses the Cross-Entropy Method (CEM)~\citep{mannor2003cross} to search for a vector that decodes into a program that maximizes the return. The rollouts of the decoded policies are used to inform the CEM search. 

\subsection{Programmatic State Machine Policies (\psm)}
\label{sec:psm}

\cite{InalaBTS20} introduced Programmatic State Machine Policies, which we refer to as \psm, a system that learns a policy as a finite-state machine. A finite state machine policy for an MDP $\mathcal{M}$ is a tuple $(M, S, A, \delta, m_0, F, \alpha) $ where $M$ is a finite set of modes. The sets $S$ and $A$ are the sets of states and actions from $\mathcal{M}$. The function $ \delta: M \times S \to M $ is the transition function, $m_0$ in $M$ is the initial mode, and $F \subseteq S$ is the set of modes in which the policy terminates. The transition function $ \delta $ defines the next mode given the current mode and input state $s$ in $S$. Finally, $\alpha : M \times S \rightarrow A$ determines the policy's action when in mode $m$ and the agent observes state $s$. 

In the \parking\ environment, \cite{InalaBTS20} considered a domain-specific language for the transition function $\delta$ and constant values for $\alpha$. The grammar defining the language $\delta$ is the following.
\begin{eqnarray*}
B & ::= & \{s[i] \geq v\}_{i=1}^n ~\vert ~ \{s[i] \leq v\}_{i=1}^n ~\vert~B \land B ~\vert~ B \lor B
\end{eqnarray*}
Here, the values $v$ are constants that need to be learned, $s[i]$ is the $i$-th entry of the state $s$ the agent observes at a given time step, and $n$ is the dimensionality of the observation. 

\begin{figure}
    \centering
\scalebox{0.8}{
\begin{tikzpicture}[node distance=6cm, square/.style={rectangle, draw, minimum size=1.3cm, line width=0.3mm, text centered}]
    % Nodes
    \node[square] (M1) {$(F, L)$};
    \node[square, left of=M1, xshift=2.5cm] (M2) {$(B, R)$};
\node[square, right of=M1, xshift=-3.25cm] (M3) {$(F, R)$};
    \node[right of=M3, xshift=-3.25cm] (Mz) {$m_z$};
    \node[above of=M1, yshift=-4.75cm] (M0) {$m_0$};

\node[above=-0.5cm of M1] {$m_1$ };
\node[above=-0.5cm of M2] {$m_2$ };
\node[above=-0.5cm of M3] {$m_3$ };
    % Arrows
    \draw[->,thick] ($(M1.north west) + (0, -0.3)$) -- ($(M2.north east) + (0, -0.3)$) node[midway, above] {$d_f \leq 0.30$};
    \draw[->,thick] (M1) -- (M3) node[midway, below] {$x < 0$};
    \draw[->,thick] ($(M2.south east) + (0, 0.3)$) -- ($(M1.south west) + (0, 0.3)$) node[midway, below] {$d_b \leq 0.30$};
    \draw[->,thick] (M3) -- (Mz) node[midway, below] {$\theta \leq 0.0$};
    \draw[->,thick] (M0) -- (M1) node[midway, below] {};
\end{tikzpicture}
}
    \caption{Example of a state machine policy, where $m_0$ is the initial mode and $m_z$ is an accepting mode. The tuples inside each mode specify the agent's action when in that mode (e.g., $(F, L)$ means ``move forward and steer to the left''. The transitions from one mode to another are triggered by a Boolean expression shown in the arrows. For example, if the car is too close to the car in front of it ($d_f \leq 0.30$), then the policy moves from $m_1$ to $m_2$. The agent remains in the current mode if no outgoing Boolean expression is triggered. This policy is based on an example by \cite{InalaBTS20}.}
    \label{fig:example_fsm}
\end{figure}
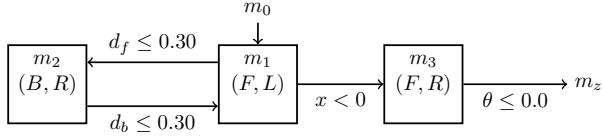

Figure~\ref{fig:example_fsm} shows an example of the type of policy \psm\ learns. In this example, the policy is for \parking, a domain where the agent must learn how to exit a parking spot with a car in front of the agent's car (car$_f$) and another at the rear (car$_b$). The policy uses the following state features: the distance between the agent's car and car$_f$ ($d_f$) and car$_b$ ($d_b$), the $x$ coordinate of the car, and the angle $\theta$ of the car. A solution involves the agent moving forward to the left (mode $m_1$) and then back to the right (mode $m_2$), until the agent has cleared car$_f$ (transitioning to mode $m_3$). 

The agent solves the problem if it straightens the car after clearing car$_f$, thus transitioning from $m_3$ to $m_f$. \psm's policies are called only once for the initial state; the policy returns only at the end of the episode. 

\psm\ learns policies with a teacher-student scheme, where the student is a finite state machine encoding the policy. The teacher is a loop-free learner that finds state-action pair sequences that optimize for two objectives. Specifically, they maximize the agent's return and minimize how much they deviate from the student's policy. Optimizing for the latter avoids sequences that cannot be encoded in a finite-state machine. After optimizing the teacher, the student updates its policy based on the teacher's sequence. The student's policy is updated through a clustering scheme on the teacher's sequence. The Boolean expressions denoting transitions between modes are found through discrete search. The process is repeated, and the teacher's sequence-based policy is optimized. 

\section{Experiments}
\label{sec:experiments}

In this section, we revisit the experiments of \cite{pirl} and \cite{verma2019imitation} on \torcs\ (Section~\ref{sec:experiments_torcs} and Appendix~\ref{sec:torcs_appendix}), of \cite{leaps} on \karel\ (Section~\ref{sec:experiments_karel} and Appendix~\ref{sec:karel_appendix}), and of \cite{InalaBTS20} on \parking\ (Section~\ref{sec:experiments_parking} and Appendix~\ref{sec:parking_appendix}). 

\subsection{\torcs}
\label{sec:experiments_torcs}

\begin{table}[t]
\centering
\scalebox{0.9}{
\begin{tabular}{lcccc}
\toprule
\textbf{}        & \textsc{Ndps}      & \textsc{Drl} ($\beta = 1.0$) & \textsc{Drl} ($\beta = 0.5$) \\
\textsc{Tracks}  & \textsc{Lap Time}  & \textsc{Lap Time}            & \textsc{Lap Time}            \\
\midrule
\textsc{G-Track-1}          & 1:01  & 54             & 1:17         \\
\textsc{G-Track-2 (OOD)}    & 1:40  & \textsc{Cr 1608m} & 1:48 (0.76)  \\
\textsc{E-Road (OOD)}       & 1:51  & \textsc{Cr 1902m} & 1:54 (0.69)  \\
\midrule
\textsc{Aalborg}            & 2:38  & 1:49           & 2:24         \\
\textsc{Alpine-2 (OOD)}     & 3:16  & \textsc{Cr 1688m} & 3:13 (1.00)  \\
\textsc{Ruudskogen (OOD)}   & 3:19  & \textsc{Cr 3232m} & 2:46 (1.00)  \\
\bottomrule
\end{tabular}
}
\vspace{4pt}
\caption{For \textsc{DRL} ($\beta = 0.5$), we trained 30 models (seeds) for \textsc{G-Track-1} and 15 for \textsc{Aalborg}. Each cell shows the average lap time (mm:ss) over three laps per model, then averaged across models; 13 models learned to complete \textsc{G-Track-1} and four models learned to complete \textsc{Aalborg}. Values in parentheses for \textsc{DRL} ($\beta=0.5$) show the fraction of seeds that successfully generalized to the test track (out of 13 and 4 for \textsc{G-Track-1} and \textsc{Aalborg}, respectively). For \ndps\ and \textsc{DRL} ($\beta = 1.0$), we used the data from \citep{pirl}, which is over three models. ``CR'' indicates that all three models crashed, and the number reported is the average distance at which the agent crashed the car.}
\label{tab:torcs}
\end{table}

\cite{pirl} and \cite{verma2019imitation} showed 
that programmatic policies written in the language from Figure~\ref{fig:torcs} generalize better to OOD problems than neural policies in race tracks of the Open Racing Car Simulator (\torcs) \citep{wymann2000torcs}. The results of \cite{pirl} also showed that neural policies better optimize the agent's return than programmatic policies, as the former complete laps more quickly than the latter on the tracks on which they are trained. We hypothesized that the programmatic policies generalize better not because of their representation, but because the car moves more slowly, thus making it easier to generalize to tracks with sharper turns. 

We test our hypothesis by training models with two different reward functions: the original function used in previous experiments ($\beta = 1.0$ in Equation~\ref{eq:reward}), which we refer to as ``original'', and a function that makes the agent more cautious about speeding ($\beta = 0.5$), which we refer to as ``cautious''. 
\begin{equation}
\beta \times V_x \cos(\theta) - \left| V_x \sin(\theta)\right| - V_x \left| d_l \right| \,.
\label{eq:reward}
\end{equation}
Here, \(V_x\) is the speed of the car along the longitudinal axis of the car, \(\theta\) is the angle between the direction of the car and the direction of the track axis, and \(d_l\) is the car’s lateral distance from the center of the track. The first term of the reward measures the velocity along the central line of the track, while the second is the velocity moving away from the central line. Maximizing the first term minus the second allows the agent to move fast without deviating from the central line. The last term also contributes to having the agent follow the center of the track. Once we set $\beta = 0.5$, the agent will learn policies where the car moves more slowly, which allows us to test our hypothesis. 

Following \cite{pirl}, we use the Deep Deterministic Policy Gradient (DDPG) algorithm \citep{lillicrap2019continuouscontroldeepreinforcement} and \torcs's practice mode, which includes 29 sensors as observation space and the actions of accelerating and steering. We considered two tracks for training the agent: \textsc{G-Track-1} and \textsc{Aalborg}. The first is considered easier than the second based on the track's number of turns, length, and width. The models trained on \textsc{G-Track-1} were tested on \textsc{G-Track-2} and \textsc{E-Road}, while the models trained on \textsc{Aalborg} were tested on \textsc{Alpine-2} and \textsc{Ruudskogen}.

Table~\ref{tab:torcs} presents the results. \ndps\ can generalize to the test problems in all three seeds evaluated. \textsc{Drl} with $\beta = 1.0$ does not generalize to the test tracks, with the numbers in the table showing the average distance at which the agent crashes the car in all three seeds. For \textsc{Drl} ($\beta = 0.5$) we trained 30 models (seeds) for \textsc{G-Track-1} and 15 for \textsc{Aalborg}. Then, we verified that 13 of the 30 models learned how to complete laps of the \textsc{G-Track-1} track, and 4 of the 15 models learned to complete laps of the \textsc{Aalborg} track; these models were evaluated on the OOD tracks. 

The results support our hypothesis that by changing the reward function, we would allow the agent to generalize. On the training tracks, the lap time increases as we reduce $\beta$. Most models trained with $\beta = 0.5$ generalize from the \textsc{G-Track-1} to \textsc{G-Track-2} (76\% of the models) and \textsc{E-Road} (69\%) tracks; all models that learned to complete a lap on \textsc{Aalborg} generalized to the other two tracks.

\subsection{\karel}
\label{sec:experiments_karel}

\begin{table}[t]
\centering
\small
\resizebox{\textwidth}{!}{%
\begin{tabular}{llccccc}
\toprule
\textbf{} & \textbf{} & \textsc{StairClimber} & \textsc{Maze} & \textsc{TopOff} & \textsc{FourCorner} & \textsc{Harvester} \\
\midrule
\multirow{2}{*}{\leaps $^\dagger$} & Small & 1.00 (0.00) & 1.00 (0.00) & 0.81 (0.07) & 0.45 (0.40) & 0.45 (0.28) \\
              & 100$\times$100  & 1.00 (0.00) & 1.00 (0.00) & 0.21 (0.03) & 0.45 (0.37) & 0.00 (0.00) \\
              \cmidrule(lr){2-7}
\multirow{2}{*}{PPO with ConvNet$^\dagger$} 
    & Small & 1.00 (0.00) & 1.00 (0.00) & 0.32 (0.07) & 0.29 (0.05) &  0.90 (0.10) \\ 
    & 100$\times$100  & 0.00 (0.00) & 0.00 (0.00) & 0.01 (0.01) & 0.00 (0.00) &  0.00 (0.00) \\
\cmidrule(lr){2-7}
\multirow{2}{*}{PPO with LSTM$^\dagger$}
    & Small & 0.13 (0.29) & 1.00 (0.00) &  0.63 (0.23) & 0.36 (0.44) &  0.32 (0.18) \\ 
    & 100$\times$100  & 0.00 (0.00) & 0.04 (0.05) & 0.15 (0.12) & 0.37 (0.44) & 0.02 (0.01) \\
    \cmidrule(lr){2-7}
\multirow{2}{*}{PPO with $a_{t-1}$} & Small & 1.00 (0.00) & 1.00 (0.00) & 1.00 (0.00) & 1.00 (0.00) & 0.59 (0.05) \\
                         & 100$\times$100  & 1.00 (0.00) & 1.00 (0.00) & 1.00 (0.00) & 1.00 (0.00) & 0.04 (0.00) \\
\bottomrule
\end{tabular}
}
\caption{Generalization results on \karel, where cells show the average return and standard deviation. ``PPO with ConvNet'' observes the entire state and employs a convolutional network to learn its representation. ``PPO with LSTM'' uses an LSTM layer for both actor and critic, while ``PPO with $a_{t-1}$'' uses a fully connected network with the observation space augmented with the agent's last action. ``Small'' refers to the problems in which the models were trained, which were of size either $8 \times 8$ or $12 \times 12$. Rows marked with a $\dagger$ are from~\cite{leaps}. The results for PPO with $a_{t-1}$ are over 30 seeds, and each seed is evaluated on 10 different initial states; the results for \leaps\ and PPO with a ConvNet and with an LSTM are over five seeds and 10 different initial states.}
\label{tab:karel_results}
\end{table}

\cite{leaps} showed that programs \leaps\ synthesized in the language shown in Figure~\ref{fig:karel_example} (a) generalized better than deep reinforcement learning baselines to problem sizes much larger than those the agent encountered during training. In our experiments, we consider the fully observable version of \karel, where the agent has access to the entire grid, and the partially observable version, where the agent can only perceive the cells around it, as shown by the non-terminal $h$ in Figure~\ref{fig:karel_example} (a). 

\begin{wrapfigure}{r}{0.475\linewidth}
    \centering›
    \includegraphics[width=0.375\linewidth]{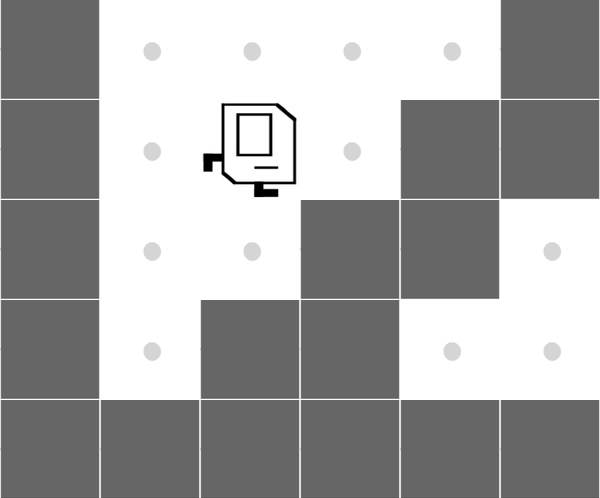}
    \includegraphics[width=0.375\linewidth]{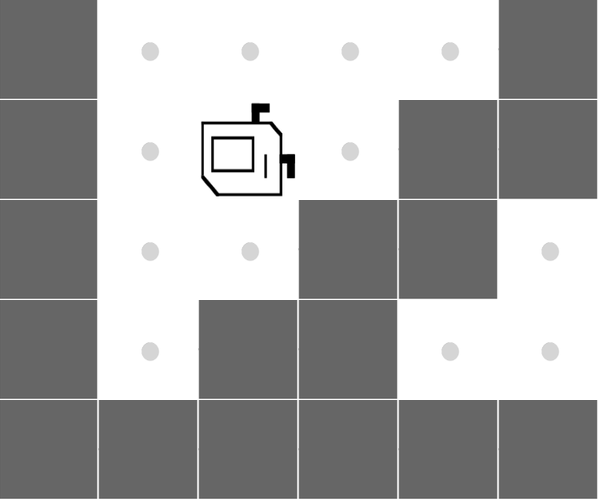}
    \caption{Different states but same observation.} \label{fig:karel_stairclimber}
\end{wrapfigure}
In the partially observable case, the problem cannot, in principle, be solved with fully connected neural networks. Consider the two states shown in Figure~\ref{fig:karel_stairclimber}. In one, the agent is going downstairs; in the other, it is going upstairs. Yet, the observation is the same for both states. \cite{leaps} used LSTMs~\citep{hochreiter1997long} to deal with the partial observability problem. Instead of using LSTMs, which tend to be more complex to train than fully connected networks, we add the last action the agent has taken as part of the observation. For the fully observable case, we report the results of \cite{leaps}, which used a convolutional network on the input. 

We trained policies for the following problems, which were chosen to match the design of \cite{leaps}: \textsc{StairClimber}, \textsc{Maze}, \textsc{TopOff}, \textsc{FourCorner}, and \textsc{Harvester}. The grid size of these problems was either $8 \times 8$ or $12 \times 12$. After learning to solve these small problems, we evaluated them on grids of size $100 \times 100$, also following \cite{leaps}. 
In the \textsc{Maze} problem, the agent learns to escape a small maze and is evaluated on a larger one. Table~\ref{tab:karel_results} shows the results.

Our results show that partial observability combined with a simpler model can generalize to larger grids. Namely, ``PPO with $a_{t-1}$'', which uses a fully connected network with the observation augmented with the agent's last action, generalizes to larger problems. This contrasts with ``PPO with ConvNet'', which operates in the fully observable setting, and ``PPO with LSTM'', which operates in the partially observable setting but uses a more complex neural model. To illustrate, in \textsc{Maze}, if the agent can only see the cells around itself, it can learn strategies such as ``follow the right wall'', which is challenging to learn in the fully observable setting. The LSTM agent fails not only to generalize to larger problems, but it often also fails to learn how to solve even the smaller problems.

\subsection{\parking}
\label{sec:experiments_parking}

\begin{table}[t]
\centering
% \small
\scalebox{0.9}{
\begin{tabular}{lcccc}
\toprule
 & \multicolumn{2}{c}{\psm} & \multicolumn{2}{c}{DQN} \\
% \cmidrule(lr){3-4} \cmidrule(lr){5-6}
 \textbf{} & Successful-on-100 & Success Rate & Successful-on-100 & Success Rate \\
\midrule
Training & 0.06 & 0.26 & 0.40 & 0.86 \\
Test     & 0.06 & 0.16 & 0.00 & 0.18 \\
\bottomrule \\
\end{tabular}
}
\caption{Evaluation of 30 seeds of \psm\ and 15 seeds of DQN on the \parking\ domain. Each model trained was evaluated on 100 different initial states of both training and testing settings. The columns ``Successful-on-100'' report the fraction of models trained that successfully solved all 100 initial states. The columns ``Success Rate'' reports the average number of initial states solved across different seeds.}
\label{tab:fsm_results}
\end{table}

In the \parking\ domain, an agent must get out of a parking spot. 
During training, the distance between the two parked cars is sampled uniformly from the range $[12.0, 13.5]$. In contrast, the test environment uses a narrower and more challenging range of $[11.0, 12.0]$, requiring the agent to generalize to tighter parking scenarios.

We evaluate both programmatic policies, as described by \citet{InalaBTS20}, and neural policies trained using Deep Q-Networks (DQN)~\citep{dqn}. Preliminary experiments showed that DQN performed better than the PPO and DDPG algorithms considered in our other experiments. For each policy type, we trained 30 independently seeded models and evaluated each one on 100 test episodes, where the test gap was sampled uniformly from the range $[11.0, 12.0]$.

Table \ref{tab:fsm_results} shows the results. We trained 30 independent models of \psm\ and 15 of DQN. Each model was evaluated on 100 different initial states. The columns ``Successful-on-100'' refer to the ratio of models that could solve all 100 initial states. For example, $0.06$ for \psm\ means that two of the 30 models solved all initial states on training and test. The ``Successful Rate'' column shows the ratio of times across all models and initial states that the learned policy could solve the problem. For example, $0.86$ for DQN in training means that DQN models solved 86\% of the $15 \times 100 = 1500$ initial states. 

Our results suggest that the \psm\ policies generalize better than the DQN policies, as two out of 30 models could solve all 100 test initial states. Looking at the difference between the ``Success Rate'' of \psm\ and DQN in training and test also suggests that \psm's policies generalize better, as the gap between the two scenarios is small for \psm: $0.26 - 0.16 = 0.10$ versus $0.86 - 0.18 = 0.68$ for DQN. However, looking at the test ``Success Rate'' alone suggests that DQN is the winner, as DQN policies can solve more test initial states on average than \psm\ can. Independent of the metric considered, our results show that \parking\ is a challenging domain for both types of representation.

\subsection{Discussion}

Our experiments showed that neural models can also generalize to OOD problems commonly used in the literature. One key aspect of programmatic solutions is the policy's sparsity. 
For example, the mode transitions in Figure~\ref{fig:example_fsm} use a single variable in the Boolean expression. By contrast, neural networks typically use all variables available while defining such transitions, often by encountering spurious correlations between input features and the agent's action. That is why providing fewer input features, combined with a simpler neural model, helped with generalization in \karel---we remove features that could generate spurious correlations with the model's actions. These results on reducing input features to enhance generalization align with other studies involving the removal of visual distractions that could hamper generalization~\citep{bertoin2022look,madi}.

In the case of \torcs, OOD generalization was possible due to a ``safer'' reward function. If the agent learns on a track that allows it to move fast and never slow down, then it is unlikely to generalize to race tracks with sharp turns that require the agent to slow down. In this case, generalization or lack thereof is not caused by the representation, but by how well the agent can optimize its return while using that representation. We conjecture that \ndps\ and \propel\ would not generalize to OOD problems if they could find better optimized policies for the agent's return in the training tracks. 

\parking\ was the most challenging benchmark we considered in our experiments, and we believe it points in the direction of benchmarks that could value the generalization power of programmatic representations. Recurrent neural networks such as LSTMs can, in principle, represent the solution shown in Figure~\ref{fig:example_fsm}. In fact, due to the loop of the agent interacting with the environment, the solution to \parking\ does not even require loops. If we augment the agent's observation with its last action, a decision tree could encode the repetitive behavior needed to solve the problem. Yet, we could not find a neural policy that reliably generalizes to OOD problems in this domain. By reliably we mean that if the agent learns how to solve the training setting, it automatically generalizes to the test setting.

\subsection{Beyond Generalization}

Our analysis has focused on generalizing to OOD problems. However, there are other important dimensions to consider when considering programmatic representations. The most common are interpretability and verifiability~\citep{viper}, as one can choose a language that results in programs that are easier for us to understand and verify.  
Intuitively, the policies of \ndps, \leaps, and \psm\ tend to be more interpretable than neural policies we learned in our experiments. 

Another important dimension is sample efficiency. A programming language's inductive bias can make the problem easier to solve. For example, we could add to the language used to define the Boolean expressions of the \psm's policies, an expression that verifies whether the agent is close to an object. Such an expression could be reused and potentially make the approach more sample-efficient. The idea of composing a solution from existing programs underlies library-learning approaches~\citep{dreamcoder,babble,stitch,aulile,decompiler}. Programmatic solutions tend to be more composable than neural ones, although recent work has investigated the decomposition of reusable pieces of neural networks~\citep{alikhasi2024unveiling}.

\section{Valuing the Generalization Power of Programmatic Policies}
\label{sec:experiments_wide_maze}

If commonly used problems undervalue the generalization power of programmatic policies, what properties of problems could showcase how programmatic policies can generalize? 
We propose an illustrative benchmark problem that requires computations that neural networks struggle to learn from data. Although recurrent models are, in theory, computationally universal~\citep{siegelmann1994analog,siegelmann1995computational}, they are more limited in practice~\citep{weiss2018practical}. We consider a problem that requires a stack or a queue, which neural models can struggle with~\citep{joulin2015inferring}.  

\begin{figure}[t]
    \centering
    \includegraphics[width=0.25\linewidth]{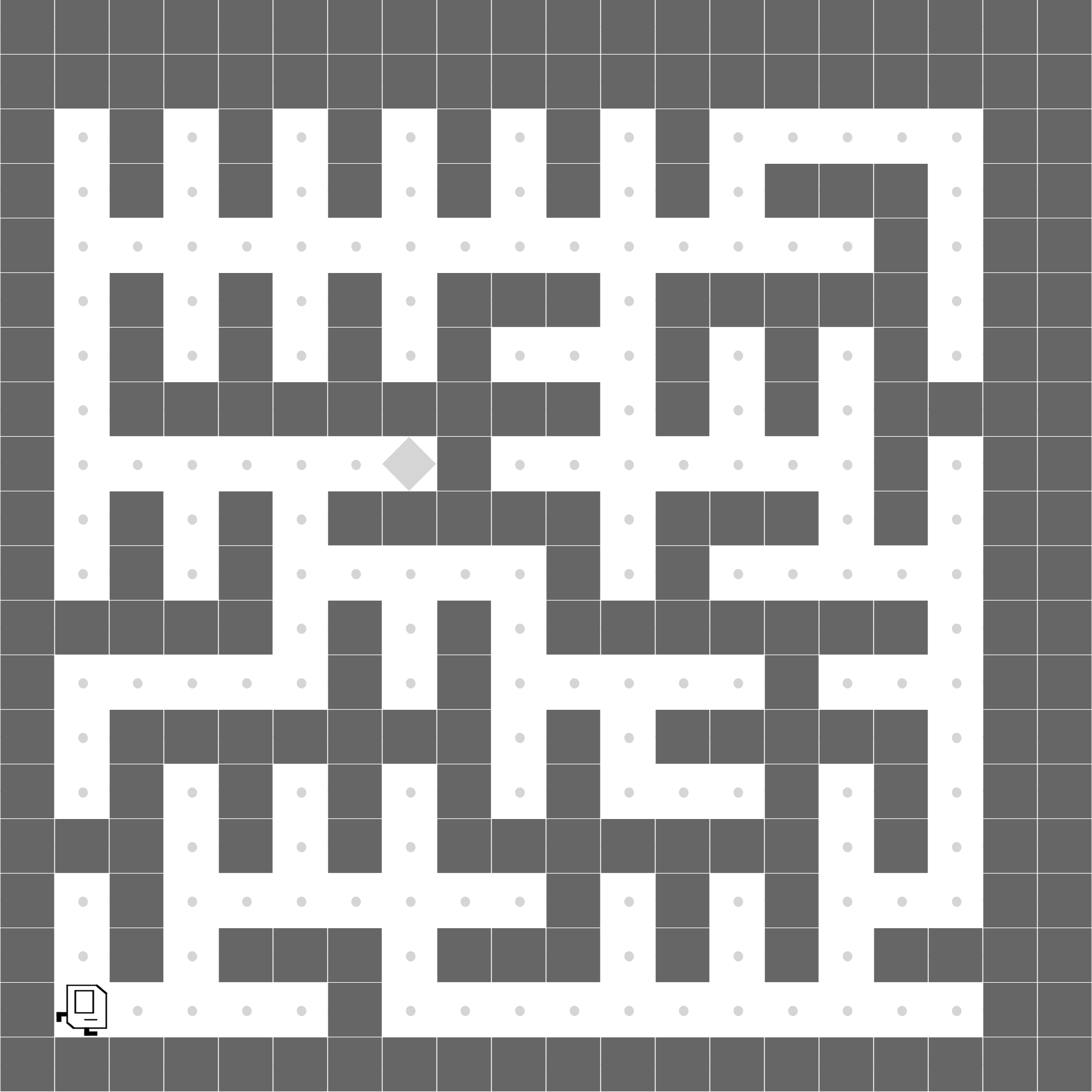}
    \includegraphics[width=0.25\linewidth]{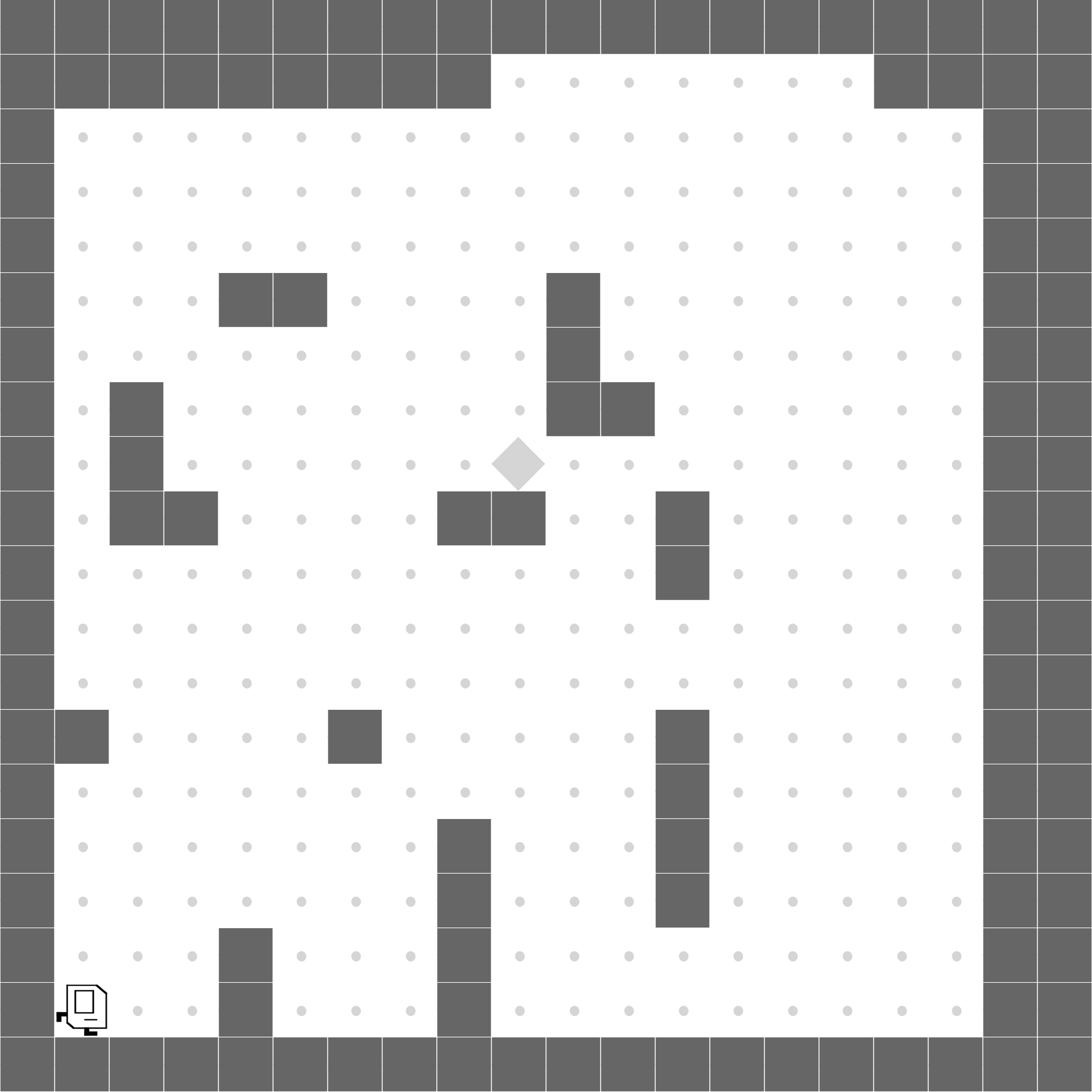}
    \caption{Maze problem from \karel\ (left) and our proposed sparse maze (right).}
    \label{fig:mazes}
\end{figure}

We consider finding the shortest paths on a grid. Suppose the agent can only sense the cells around itself, as in the \karel\ problem. If the environment is not dense in walls, such that the agent can use simple strategies such as ``follow the right wall'', it needs to remember the cells it has visited to find the shortest path from its initial location to a goal location. Iterative-Deepening Depth-First Search (IDDFS) uses a stack to solve shortest-path problems. Dijkstra's algorithm~\citep{dijkstra1959note} could also be used, but it requires that the agent ``jumps around'' the state space as states far from each other can be expanded from one time step to the next based on the priority of the algorithm's queue. 

The maze environment from the \karel\ benchmark is similar to the problem we consider, which we call SparseMaze; see Figure~\ref{fig:mazes}. What makes the \karel\ mazes easier than what we propose is that the agent always has a wall as a reference, favoring strategies such as ``follow the right wall'' that do not require memory use. If the map is sparse, as in Figure~\ref{fig:mazes} (right), finding any solution, let alone finding the shortest one, becomes challenging due to the model's inability of learning stacks and queues. 

Table~\ref{tab:sparse_maze} presents the generalization results of neural and programmatic policies in \textsc{SparseMaze}. As neural policies, we considered PPO with a GRU~\citep{chung2014empiricalevaluationgatedrecurrent}. As for a programmatic policy, we used \textsc{FunSearch} \citep{FunSearch2023} with Qwen 2.5-Coder (32B) \citep{bai2023qwentechnicalreport} to synthesize Python programs encoding policies to \textsc{SparseMaze}. We use the return of a rollout of the policies as the evaluation function in \textsc{FunSearch}. Appendix~\ref{sec:sparse_maze} provides the training information of the neural policies and the prompt used in \textsc{FunSearch}. Each approach was trained on maps of size $20 \times 20$ (``Original'' in Table~\ref{tab:sparse_maze}), and evaluated on maps of size $100 \times 100$. While PPO could not learn a good policy even for the smaller map, \textsc{FunSearch} synthesized the breadth-first search (BFS) algorithm after 21 iterations of evolution, which generalizes to maps of any size (see Appendix~\ref{sec:sparse_maze} for \textsc{FunSearch}'s policy). Similarly to Dijkstra's algorithm, BFS also uses a queue and thus assumes that the agent can ``jump around'' the state space. Nevertheless, this proof-of-concept experiment shows an example where programmatic representations can generalize to OOD problems, while neural policies are unlikely to generalize. 

\begin{table}[t]
\centering
\begin{tabular}{llc}
\toprule
Model & \textbf{} & Return (std)\\
\midrule
\textsc{FunSearch} & Original & \textbf{1.00 (0.00)} \\
                        & $100 \times 100$ & \textbf{1.00 (0.00)} \\
PPO with GRU & Original & 0.09 (0.29) \\
             & $100 \times 100$ & 0.11 (0.32) \\
\bottomrule
\end{tabular}
\vspace{4pt}
\caption{Results on \textsc{SparseMaze}. The return is an average over 10 initial states. Results for PPO are averaged over 30 seeds; results for \textsc{FunSearch} are a single run of the system.}
\label{tab:sparse_maze}
\end{table}

\section{Conclusion}

In this paper, we argued that commonly used benchmarks undervalue the generalization capabilities of programmatic representations. We empirically showed that the OOD generalization gap between programmatic and neural policies on commonly used benchmarks is not as large as we previously thought. We showed that simple neural networks can generalize to OOD problems in \karel\ problems when using the sparse observations of programmatic representations. We also showed that, due to neural policies' ability to optimize the agent's return, they might become too specialized in a problem to generalize OOD. This can be fixed in \torcs\ by using a more cautious reward function that allows the car to move more slowly and thus generalize to unseen and possibly more challenging race tracks. We also evaluated the \parking\ problem, where the agent must learn repetitive behaviors. While programmatic policies generalized slightly better than neural policies in this domain, both representations struggled to learn policies that generalize reliably, thus suggesting that more research is needed to understand what is required to attain OOD generalization in this type of problem. Finally, we argued for benchmarks focusing on the weaknesses of neural networks. As an illustrative example, we suggested a benchmark that requires the agent to use data structures that neural networks struggle to learn. By focusing on the weaknesses of neural networks, we will better understand the scenarios in which programmatic representations are needed and, importantly, how programmatic and neural representations can be combined into representations that inherit the best properties of both worlds.  

\bibliographystyle{named}
\bibliography{references}

\begin{thebibliography}{}

\bibitem[\protect\citeauthoryear{Albarghouthi \bgroup \em et al.\egroup }{2013}]{AlbarghouthiGK13}
Aws Albarghouthi, Sumit Gulwani, and Zachary Kincaid.
\newblock Recursive program synthesis.
\newblock In {\em International Conference Computer Aided Verification, {CAV}}, pages 934--950, 2013.

\bibitem[\protect\citeauthoryear{Alikhasi and Lelis}{2024}]{alikhasi2024unveiling}
Mahdi Alikhasi and Levi Lelis.
\newblock Unveiling options with neural network decomposition.
\newblock In {\em The Twelfth International Conference on Learning Representations}, 2024.

\bibitem[\protect\citeauthoryear{Bai \bgroup \em et al.\egroup }{2023}]{bai2023qwentechnicalreport}
Jinze Bai, Shuai Bai, Yunfei Chu, Zeyu Cui, Kai Dang, Xiaodong Deng, Yang Fan, Wenbin Ge, Yu~Han, Fei Huang, Binyuan Hui, Luo Ji, Mei Li, Junyang Lin, Runji Lin, Dayiheng Liu, Gao Liu, Chengqiang Lu, Keming Lu, Jianxin Ma, Rui Men, Xingzhang Ren, Xuancheng Ren, Chuanqi Tan, Sinan Tan, Jianhong Tu, Peng Wang, Shijie Wang, Wei Wang, Shengguang Wu, Benfeng Xu, Jin Xu, An~Yang, Hao Yang, Jian Yang, Shusheng Yang, Yang Yao, Bowen Yu, Hongyi Yuan, Zheng Yuan, Jianwei Zhang, Xingxuan Zhang, Yichang Zhang, Zhenru Zhang, Chang Zhou, Jingren Zhou, Xiaohuan Zhou, and Tianhang Zhu.
\newblock Qwen technical report, 2023.

\bibitem[\protect\citeauthoryear{Bastani \bgroup \em et al.\egroup }{2018}]{viper}
Osbert Bastani, Yewen Pu, and Armando Solar-Lezama.
\newblock Verifiable reinforcement learning via policy extraction.
\newblock In {\em Proceedings of the International Conference on Neural Information Processing Systems}, pages 2499--2509, 2018.

\bibitem[\protect\citeauthoryear{Bertoin \bgroup \em et al.\egroup }{2022}]{bertoin2022look}
David Bertoin, Adil Zouitine, Mehdi Zouitine, and Emmanuel Rachelson.
\newblock Look where you look! saliency-guided q-networks for generalization in visual reinforcement learning.
\newblock In Alice~H. Oh, Alekh Agarwal, Danielle Belgrave, and Kyunghyun Cho, editors, {\em Advances in Neural Information Processing Systems}, 2022.

\bibitem[\protect\citeauthoryear{Bowers \bgroup \em et al.\egroup }{2023}]{stitch}
Matthew Bowers, Theo~X. Olausson, Lionel Wong, Gabriel Grand, Joshua~B. Tenenbaum, Kevin Ellis, and Armando Solar-Lezama.
\newblock Top-down synthesis for library learning.
\newblock {\em Proceedings of the ACM on Programming Languages}, 2023.

\bibitem[\protect\citeauthoryear{Cao \bgroup \em et al.\egroup }{2023}]{babble}
David Cao, Rose Kunkel, Chandrakana Nandi, Max Willsey, Zachary Tatlock, and Nadia Polikarpova.
\newblock Babble: Learning better abstractions with e-graphs and anti-unification.
\newblock {\em Proceedings of the ACM on Programming Languages}, 2023.

\bibitem[\protect\citeauthoryear{Chung \bgroup \em et al.\egroup }{2014}]{chung2014empiricalevaluationgatedrecurrent}
Junyoung Chung, Caglar Gulcehre, KyungHyun Cho, and Yoshua Bengio.
\newblock Empirical evaluation of gated recurrent neural networks on sequence modeling, 2014.

\bibitem[\protect\citeauthoryear{Dijkstra}{1959}]{dijkstra1959note}
Edsger~W. Dijkstra.
\newblock A note on two problems in connexion with graphs.
\newblock {\em Numerische Mathematik}, 1:269--271, 1959.

\bibitem[\protect\citeauthoryear{Ellis \bgroup \em et al.\egroup }{2023}]{dreamcoder}
Kevin Ellis, Lionel Wong, Maxwell Nye, Mathias Sabl-Meyer, Luc Cary, Lore Pozo, Luke Hewitt, Armando Solar-Lezama, and Joshua Tenenbaum.
\newblock Dreamcoder: growing generalizable, interpretable knowledge with wake?sleep bayesian program learning.
\newblock {\em Philosophical Transactions of the Royal Society A: Mathematical, Physical and Engineering Sciences}, 381, 06 2023.

\bibitem[\protect\citeauthoryear{Grooten \bgroup \em et al.\egroup }{2024}]{madi}
Bram Grooten, Tristan Tomilin, Gautham Vasan, Matthew~E. Taylor, A.~Rupam Mahmood, Meng Fang, Mykola Pechenizkiy, and Decebal~Constantin Mocanu.
\newblock Madi: Learning to mask distractions for generalization in visual deep reinforcement learning.
\newblock In {\em Proceedings of the International Conference on Autonomous Agents and Multiagent Systems}, page 733?742. International Foundation for Autonomous Agents and Multiagent Systems, 2024.

\bibitem[\protect\citeauthoryear{Hochreiter and Schmidhuber}{1997}]{hochreiter1997long}
Sepp Hochreiter and J{\"u}rgen Schmidhuber.
\newblock Long short-term memory.
\newblock {\em Neural Computation}, 9(8):1735--1780, 1997.

\bibitem[\protect\citeauthoryear{Inala \bgroup \em et al.\egroup }{2020}]{InalaBTS20}
Jeevana~Priya Inala, Osbert Bastani, Zenna Tavares, and Armando Solar{-}Lezama.
\newblock Synthesizing programmatic policies that inductively generalize.
\newblock In {\em International Conference on Learning Representations}, 2020.

\bibitem[\protect\citeauthoryear{Joulin and Mikolov}{2015}]{joulin2015inferring}
Armand Joulin and Tomas Mikolov.
\newblock Inferring algorithmic patterns with stack-augmented recurrent nets.
\newblock In {\em Advances in Neural Information Processing Systems (NeurIPS)}, volume~28, 2015.

\bibitem[\protect\citeauthoryear{Kingma and Welling}{2014}]{kingma2014autoencoding}
Diederik~P. Kingma and Max Welling.
\newblock Auto-encoding variational bayes.
\newblock In {\em International Conference on Learning Representations}, 2014.

\bibitem[\protect\citeauthoryear{Lillicrap \bgroup \em et al.\egroup }{2019}]{lillicrap2019continuouscontroldeepreinforcement}
Timothy~P. Lillicrap, Jonathan~J. Hunt, Alexander Pritzel, Nicolas Heess, Tom Erez, Yuval Tassa, David Silver, and Daan Wierstra.
\newblock Continuous control with deep reinforcement learning, 2019.

\bibitem[\protect\citeauthoryear{Mannor \bgroup \em et al.\egroup }{2003}]{mannor2003cross}
Shie Mannor, Reuven~Y. Rubinstein, and Yohai Gat.
\newblock The cross entropy method for fast policy search.
\newblock In {\em Proceedings of the International Conference on Machine Learning}, pages 512--519, 2003.

\bibitem[\protect\citeauthoryear{Mnih \bgroup \em et al.\egroup }{2015}]{dqn}
Volodymyr Mnih, Koray Kavukcuoglu, David Silver, Andrei~A Rusu, Joel Veness, Marc~G Bellemare, Alex Graves, Martin Riedmiller, Andreas~K Fidjeland, Georg Ostrovski, et~al.
\newblock Human-level control through deep reinforcement learning.
\newblock {\em Nature}, 518(7540):529--533, 2015.

\bibitem[\protect\citeauthoryear{Palmarini \bgroup \em et al.\egroup }{2024}]{decompiler}
Alessandro~B. Palmarini, Christopher~G. Lucas, and N.~Siddharth.
\newblock Bayesian program learning by decompiling amortized knowledge.
\newblock In {\em Proceedings of the International Conference on Machine Learning}, 2024.

\bibitem[\protect\citeauthoryear{Qiu and Zhu}{2021}]{qiu2021programmatic}
Wenjie Qiu and He~Zhu.
\newblock Programmatic reinforcement learning without oracles.
\newblock In {\em International Conference on Learning Representations}, 2021.

\bibitem[\protect\citeauthoryear{Rahman \bgroup \em et al.\egroup }{2024}]{aulile}
Habibur Rahman, Thirupathi~Reddy Emireddy, Kenneth Tjhia, Elham Parhizkar, and Levi Lelis.
\newblock Synthesizing libraries of programs with auxiliary functions.
\newblock {\em Transactions on Machine Learning Research}, 2024.

\bibitem[\protect\citeauthoryear{Romera-Paredes \bgroup \em et al.\egroup }{2023}]{FunSearch2023}
Bernardino Romera-Paredes, Mohammadamin Barekatain, Alexander Novikov, Matej Balog, M.~Pawan Kumar, Emilien Dupont, Francisco J.~R. Ruiz, Jordan Ellenberg, Pengming Wang, Omar Fawzi, Pushmeet Kohli, and Alhussein Fawzi.
\newblock Mathematical discoveries from program search with large language models.
\newblock {\em Nature}, 2023.

\bibitem[\protect\citeauthoryear{Ross \bgroup \em et al.\egroup }{2011}]{ross2011reduction}
Stéphane Ross, Geoffrey Gordon, and Drew Bagnell.
\newblock A reduction of imitation learning and structured prediction to no-regret online learning.
\newblock In {\em Proceedings of the Fourteenth International Conference on Artificial Intelligence and Statistics}, pages 627--635. PMLR, 2011.

\bibitem[\protect\citeauthoryear{Schulman \bgroup \em et al.\egroup }{2017}]{schulman2017proximal}
John Schulman, Filip Wolski, Prafulla Dhariwal, Alec Radford, and Oleg Klimov.
\newblock Proximal policy optimization algorithms.
\newblock {\em arXiv preprint arXiv:1707.06347}, 2017.

\bibitem[\protect\citeauthoryear{Siegelmann and Sontag}{1994}]{siegelmann1994analog}
Hava~T. Siegelmann and Eduardo~D. Sontag.
\newblock Analog computation via neural networks.
\newblock {\em Theoretical Computer Science}, 131(2):331--360, 1994.

\bibitem[\protect\citeauthoryear{Siegelmann and Sontag}{1995}]{siegelmann1995computational}
Hava~T. Siegelmann and Eduardo~D. Sontag.
\newblock On the computational power of neural nets.
\newblock {\em Journal of Computer and System Sciences}, 50(1):132--150, 1995.

\bibitem[\protect\citeauthoryear{Snoek \bgroup \em et al.\egroup }{2012}]{snoek2012practical}
Jasper Snoek, Hugo Larochelle, and Ryan~P. Adams.
\newblock Practical bayesian optimization of machine learning algorithms.
\newblock In {\em Advances in Neural Information Processing Systems}, volume~25, pages 2951--2959, 2012.

\bibitem[\protect\citeauthoryear{Trivedi \bgroup \em et al.\egroup }{2021}]{leaps}
Dweep Trivedi, Jesse Zhang, Shao{-}Hua Sun, and Joseph~J. Lim.
\newblock Learning to synthesize programs as interpretable and generalizable policies.
\newblock In {\em Advances in Neural Information Processing Systems}, pages 25146--25163, 2021.

\bibitem[\protect\citeauthoryear{Udupa \bgroup \em et al.\egroup }{2013}]{Udupa:2013}
Abhishek Udupa, Arun Raghavan, Jyotirmoy~V. Deshmukh, Sela Mador-Haim, Milo~M.K. Martin, and Rajeev Alur.
\newblock Transit: Specifying protocols with concolic snippets.
\newblock In {\em Proceedings of the ACM SIGPLAN Conference on Programming Language Design and Implementation}, pages 287--296. ACM, 2013.

\bibitem[\protect\citeauthoryear{Verma \bgroup \em et al.\egroup }{2018}]{pirl}
Abhinav Verma, Vijayaraghavan Murali, Rishabh Singh, Pushmeet Kohli, and Swarat Chaudhuri.
\newblock Programmatically interpretable reinforcement learning.
\newblock In {\em Proceedings of the International Conference on Machine Learning}, volume~80, pages 5045--5054. PMLR, 2018.

\bibitem[\protect\citeauthoryear{Verma \bgroup \em et al.\egroup }{2019}]{verma2019imitation}
Abhinav Verma, Hoang Le, Yisong Yue, and Swarat Chaudhuri.
\newblock Imitation-projected programmatic reinforcement learning.
\newblock {\em Advances in Neural Information Processing Systems}, 32, 2019.

\bibitem[\protect\citeauthoryear{Weiss \bgroup \em et al.\egroup }{2018}]{weiss2018practical}
Gail Weiss, Yoav Goldberg, and Eran Yahav.
\newblock On the practical computational power of finite precision rnns for language recognition.
\newblock In {\em Proceedings of the Annual Meeting of the Association for Computational Linguistics}, pages 740--745. Association for Computational Linguistics, 2018.

\bibitem[\protect\citeauthoryear{Wymann \bgroup \em et al.\egroup }{2000}]{wymann2000torcs}
Bernhard Wymann, Eric Espi{\'e}, Christophe Guionneau, Christos Dimitrakakis, R{\'e}mi Coulom, and Andrew Sumner.
\newblock Torcs, the open racing car simulator.
\newblock {\em Software available at http://torcs. sourceforge. net}, 4(6):2, 2000.

\end{thebibliography}

\clearpage
\appendix

\section{\textsc{TORCS} Details}
\label{sec:torcs_appendix}

We use the hyperparameters in Table~\ref{tab:ddpg_config} with DDPG \citep{lillicrap2019continuouscontroldeepreinforcement}. 

\begin{table}[H]
\centering
\begin{tabular}{lc}
\toprule
\textbf{Hyperparameter} & \textbf{Selected Value} \\
\midrule
Actor's learning rate           & \texttt{0.0003} \\
Critic's learning rate           & \texttt{0.001} \\
Batch size              & \texttt{64} \\
Buffer size              & \texttt{100000} \\
$\tau$              & \texttt{0.005} \\
L1 regularization       & \texttt{0.00001} \\
% Actor's hidden layer size       & \texttt{32} \\
Max steps              & \texttt{20000} \\
Training episodes      & \texttt{600} \\
\bottomrule
\end{tabular}
\vspace{4pt}
\caption{Hyperparameter Configuration Used for TORCS}
\label{tab:ddpg_config}
\end{table}

\begin{figure}[ht]
    \centering
    \includegraphics[width=0.47\linewidth, angle=90]{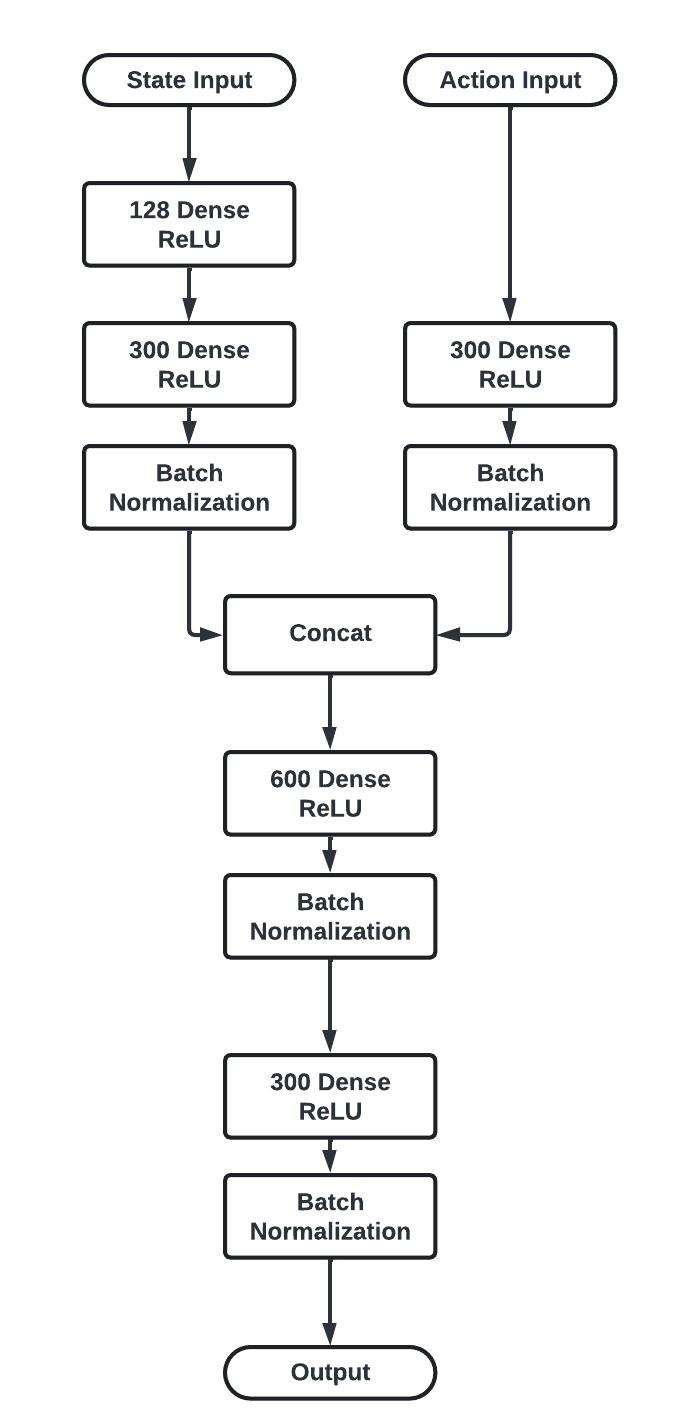}
    \caption{Architecture of the critic network used in DDPG for the TORCS environment.}
    \label{fig:ddpg_critic}
\end{figure}

\section{\textsc{Karel} Details}
\label{sec:karel_appendix}

We used Proximal Policy Optimization (PPO)~\citep{schulman2017proximal} with the agent’s previous action appended to the observation vector. A comprehensive hyperparameter sweep was conducted over the values from Table~\ref{tab:ppo_sweep}.

\begin{table}[ht]
\centering
\begin{tabular}{lc}
\toprule
\textbf{Hyperparameter} & \textbf{Values Tested} \\
\midrule
Learning rate          & \{\texttt{0.001}, \texttt{0.0001}, \texttt{0.00001}\} \\
Clipping coefficient   & \{\texttt{0.01}, \texttt{0.1}, \texttt{0.2}\} \\
Entropy coefficient    & \{\texttt{0.001}, \texttt{0.01}, \texttt{0.1}\} \\
L1 regularization      & \{\texttt{0.0}, \texttt{0.0001}, \texttt{0.0005}, \texttt{0.001}\} \\
Actor's hidden layer size      & \{\texttt{32}, \texttt{64}\} \\
Training time steps    & \{\texttt{2 million}\} \\
Seeds per config       & \{\texttt{3}\} \\
\bottomrule
\end{tabular}
\vspace{4pt}
\caption{PPO + Last Action: Hyperparameter Sweep Configuration for Karel}
\label{tab:ppo_sweep}
\end{table}

The max steps used for training and testing different Karel tasks are shown in Table~\ref{tab:karel_maxsteps}.
\begin{table}[ht]
\centering
\begin{tabular}{lccccc}
         & \textsc{StairClimber} & \textsc{Maze} & \textsc{TopOff} & \textsc{FourCorner} & \textsc{Harvester} \\
         \midrule
\textsc{Training} & 50   &  100    & 100  & 100  & 200     \\
\textsc{Test}     & 1000 &  100000 & 1000 & 1000 &  10000  \\
\bottomrule
\end{tabular}
\vspace{4pt}
\caption{Max steps of episodes for each Karel task during training and test.}
\label{tab:karel_maxsteps}
\end{table}

Table~\ref{tab:ppo_best_config} shows the best-performing configuration across all five tasks for final evaluation. The selection was based on the agent's average return across the three seeds after two million time steps of training.

\begin{table}[H]
\centering
\begin{tabular}{lc}
\toprule
\textbf{Hyperparameter} & \textbf{Selected Value} \\
\midrule
Learning rate           & \texttt{0.001} \\
Clipping coefficient    & \texttt{0.1} \\
Entropy coefficient     & \texttt{0.1} \\
L1 regularization       & \texttt{0.0} \\
Actor's hidden layer size       & \texttt{32} \\
\bottomrule
\end{tabular}
\vspace{4pt}
\caption{Best Hyperparameter Configuration Used for Final Training of Karel}
\label{tab:ppo_best_config}
\end{table}

The best configuration was then trained with 30 random seeds, and evaluation results were averaged over 10 distinct initial configurations per seed. Additionally, four other hyperparameter configurations achieved 100\% generalization on four out of five tasks: \textsc{StairClimber}, \textsc{Maze}, and \textsc{TopOff}.

Training used diverse initial state configurations. Whenever feasible, we enumerated all combinations of agent and goal placements. Specifically:
\begin{itemize}
    \item For \textsc{StairClimber}, \textsc{TopOff}, and \textsc{FourCorner}, all possible agent-goal placements were used.
    \item For \textsc{Maze}, where full enumeration was computationally infeasible, we sampled 5 random mazes and placed the goal at every position on the grid.
\end{itemize}

The training grid sizes for each task were:
\begin{itemize}
    \item $12 \times 12$ for \textsc{StairClimber}, \textsc{TopOff}, and \textsc{FourCorner}
    \item $8 \times 8$ for \textsc{Maze} and \textsc{Harvester}
\end{itemize}

\section{\parking\ Details}
\label{sec:parking_appendix}

We used a single-hidden-layer DQN architecture with 64 units for the neural baseline. The agent operated over a discretized action space, where continuous actions were mapped onto $n$ equally spaced values using a fixed action resolution. 
We performed a grid search over the hyperparameter values listed in Table \ref{tab:dqn_sweep}.
The selected hyper-parameters are shown in Table \ref{tab:dqn_best_config}.
\begin{table}[ht]
\centering
\begin{tabular}{lc}
\toprule
\textbf{Hyperparameter} & \textbf{Values Tested} \\
\midrule
Learning rate          & \{\texttt{0.01}, \texttt{0.001}, \texttt{0.0001}\} \\
Batch size   & \{\texttt{64}, \texttt{128}, \texttt{256}\} \\
Target update frequency    & \{\texttt{100}, \texttt{500}, \texttt{1000}\} \\
$\epsilon$      & \{\texttt{0.1}, \texttt{0.01}\} \\
Replay buffer size      & \{\texttt{1 million}, \texttt{2 million}\} \\
Action resolution    & \{\texttt{3}, \texttt{5}, \texttt{7}\} \\
Seeds per config       & \{\texttt{10}\} \\
\bottomrule
\end{tabular}
\vspace{4pt}
\caption{DQN: Hyperparameter Sweep Configuration for Parking Domain}
\label{tab:dqn_sweep}
\end{table}

\begin{table}[H]
\centering
\begin{tabular}{lc}
\toprule
\textbf{Hyperparameter} & \textbf{Selected Value} \\
\midrule
Learning rate           & \texttt{0.0001} \\
Batch size    & \texttt{64} \\
Target update frequency     & \texttt{1000} \\
$\epsilon$        & \texttt{0.01} \\
Action resolution      & \texttt{2 million} \\
\bottomrule
\end{tabular}
\vspace{4pt}
\caption{Best Hyperparameter Configuration Used for Final Training}
\label{tab:dqn_best_config}
\end{table}

The original \parking\ benchmark introduced by \citet{InalaBTS20} was not designed with reinforcement learning in mind—it provides ``safety check'' to invalidate policies that crash the car or get out of boundaries. 
We define both a shaped reward function and a termination condition to adapt it for RL. 
If the agent successfully reaches the parking exit, the episode ends with a large positive reward (2 × max episode length); if it takes an unsafe action, it terminates immediately with a large negative penalty (–2 × max episode length). Otherwise, at each timestep the agent receives
$r_t = - \bigl(2 \lvert x_{\text{agent}} - x_{\text{goal}}\bigr\rvert + \lvert y_{\text{agent}} - y_{\text{goal}}\bigr\rvert\bigr) - 1$,
i.e., the (weighted) negative Manhattan distance minus an extra step penalty of 1, encouraging the car to move closer to the exit.

\section{\textsc{SparseMaze} Details}
\label{sec:sparse_maze}

We trained agents using Proximal Policy Optimization (PPO)~\citep{schulman2017proximal} with the previous action included in the observation vector. A hyperparameter sweep was conducted with the values in Table~\ref{tab:wide_maze_sweep}.

\begin{table}[ht]
\centering
\begin{tabular}{lc}
\toprule
\textbf{Hyperparameter} & \textbf{Values Tested} \\
\midrule
Learning rate           & \{\texttt{0.001}, \texttt{0.0001}, \texttt{0.00001}\} \\
Clipping coefficient    & \{\texttt{0.01}, \texttt{0.1}, \texttt{0.2}\} \\
Entropy coefficient     & \{\texttt{0.001}, \texttt{0.01}, \texttt{0.1}\} \\
L1 regularization       & \{\texttt{0.0}, \texttt{0.0001}, \texttt{0.0005}, \texttt{0.001}\} \\
Actor's hidden layer size       & \{\texttt{32}, \texttt{64}\} \\
Number of minibatches   & \{\texttt{32}, \texttt{64}, \texttt{128}\} \\
Training time steps     & \{\texttt{5 million}\} \\
Seeds per config & \{\texttt{5}\} \\
\bottomrule
\end{tabular}
\vspace{4pt}
\caption{PPO with $a_{t-1}$ and GRU: Hyperparameter Sweep for \textsc{SparseMaze}}
\label{tab:wide_maze_sweep}
\end{table}

As shown in Table~\ref{tab:wide_maze_sweep}, we evaluated a range of hyperparameters, and selected the best set listed in Table~\ref{tab:wide_maze_best} for PPO with $a_{t-1}$ and in Table~\ref{tab:wide_maze_best_gru} for PPO with GRU based on training AUC. This configuration was then trained with 30 seeds, and evaluation results were averaged over 10 distinct environment seeds per model. The environment grid size was set to $20 \times 20$ during training. A smaller grid, such as $8 \times 8$, would make the maze too sparse for effective learning in this domain. Results are available in Table~\ref{tab:sparse_maze}.

\begin{table}[ht]
\centering
\begin{tabular}{lc}
\toprule
\textbf{Hyperparameter} & \textbf{Selected Value} \\
\midrule
Learning rate     & \texttt{0.0001} \\
Actor's hidden layer size       & \texttt{64} \\
Entropy coefficient     & \texttt{0.01} \\
Clipping coefficient    & \texttt{0.1} \\
Number of minibatches   & \texttt{32} \\
\bottomrule
\end{tabular}
\vspace{4pt}
\caption{Best Hyperparameter Configuration of PPO with $a_{t-1}$ for \textsc{SparseMaze}}
\label{tab:wide_maze_best}
\end{table}

\begin{table}[ht]
\centering
\begin{tabular}{lc}
\toprule
\textbf{Hyperparameter} & \textbf{Selected Value} \\
\midrule
Learning rate     & \texttt{0.0001} \\
Actor's hidden layer size       & \texttt{32} \\
Entropy coefficient     & \texttt{0.01} \\
Clipping coefficient    & \texttt{0.2} \\
Number of minibatches   & \texttt{32} \\
Value learning rate     & \texttt{0.0005} \\
GRU hidden layer size   & \texttt{64} \\
\bottomrule
\end{tabular}
\vspace{4pt}
\caption{Best Hyperparameter Configuration of PPO with GRU for \textsc{SparseMaze}}
\label{tab:wide_maze_best_gru}
\end{table}

% Funsearch Policy
Listing~\ref{lst:funsearch_policy} shows the policy that Funsearch \citep{FunSearch2023} gave after 21 iterations and Listing~\ref{lst:funsearch_prompt} is the given prompt. We used Qwen 2.5-Coder, 32B variant \citep{bai2023qwentechnicalreport} as the LLM for this part. 

\begin{lstlisting}[language=Python, caption=Funsearch Policy, captionpos=t, label={lst:funsearch_policy}]
def get_action(env: KarelGymEnv) -> Union[list[int], str]:
  """Creates a policy that returns a list of actions for the Karel agent to take in the environment."""
  
  from collections import deque
  
  def find_path():
    # Breadth-first search (BFS) to find a path to the goal
    r, c, d = env.task.get_hero_pos()
    queue = deque([(r, c, [])])
    visited = set((r, c))
    
    state_arr = env.task.get_state()
    walls = state_arr[4].astype(bool)
    directions = [(0, 1), (1, 0), (0, -1), (-1, 0)]
    
    while queue:
      current_r, current_c, path = queue.popleft()
      
      if (current_r, current_c) == env.task_specific.marker_position:
        return path
      
      for dr, dc in directions:
        new_r, new_c = current_r + dr, current_c + dc
        if 0 <= new_r < env.env_height and 0 <= new_c < env.env_width and not walls[new_r, new_c] and (new_r, new_c) not in visited:
          visited.add((new_r, new_c))
          queue.append((new_r, new_c, path + [(dr, dc)]))
    
    return []
  
  def convert_path_to_actions(path):
    actions = []
    current_r, current_c, d = env.task.get_hero_pos()
    directions_map = {(0, 1): (1, 2), (1, 0): (3, 4), (0, -1): (3, 1), (-1, 0): (4, 2)}
    
    for dr, dc in path:
      target_direction = directions_map[(dr, dc)]
      
      while d not in target_direction:
        if (d + 1) % 4 == target_direction[1]:  # Turn right
          actions.append(2)
        elif (d - 1) % 4 == target_direction[0]:  # Turn left
          actions.append(1)
        else:  # Turn around
          actions.extend([1, 1])
        
        d = (d + 1) % 4 if d in [target_direction[0], target_direction[1]] else (d - 1) % 4
      
      actions.append(0)  # Move forward
    
    return actions
  
  path = find_path()
  actions = convert_path_to_actions(path)
  
  if not actions:
    actions = [random.randint(0, 4) for _ in range(50)]
  
  return actions
\end{lstlisting}

\begin{lstlisting}[language=Python, caption=Funsearch Prompt, captionpos=t, label={lst:funsearch_prompt}]
"""
Specification for the Karel SparseMaze environment.

We are searching for a function `get_action(env)` that returns a list of actions list[int] for the Karel environment.
get_action(env) should return a policy that can solve the maze all the time, regardless of the initial configuration. Then by calling this policy, it can get the actions for that specific initial configuration.

Input is a KarelGymEnv object.
- You can access the height and width of the env like this: env.env_width, env.env_height
- You can access walls like this:
  # static walls from feature index 4 of the Karel state
  state_arr = env.task.get_state()             # shape: (features, H, W)
  self.walls = state_arr[4].astype(bool)       # True where wall
- You can the row, column, and direction of the agenr like this:
  r, c, d = env.task.get_hero_pos()
- And the directions are like this:
  0: 'Karel facing North',
  1: 'Karel facing East',
  2: 'Karel facing South',
  3: 'Karel facing West',
- Access the goal marker position like this:
  goal_r, goal_c = env.task_specific.marker_position
- You can access the observation like this:
  obs = env._get_observation_dsl()  # shape: (4,), [frontIsClear, leftIsClear, rightIsClear, markersPresent]

The actions are:
  0: move
  1: turnLeft
  2: turnRight
  3: pickMarker (not used in maze)
  4: putMarker (not used in maze)

In the maze task, the agent starts at a fixed position and must find its path to a goal marker. The environment uses a sparse reward: 1 when reaching the goal, 0 otherwise.
The environment is a sparse maze (corridors are 2 cells wide) and has multiple initial configurations (both mazes and goal positions).

This specification describes the key classes, variables, and functions used to define a Gym compatible "Karel SparseMaze" task, where an agent navigates a 
carved maze to reach a goal marker under sparse rewards.

Package Layout:
  project_root/
  |-- funsearch/
      |-- implementation/
      |   |-- utils.py     
      |   |-- temp1.py      # top-level script that calls evaluate and get_action
      |-- karel_wide_maze/
      |   |-- __init__.py
      |   |-- karel_wide_maze.py
      |   |-- karel_wide_maze_prompt_spec.py
      |   |-- gym_envs/
      |   |   |-- __init__.py
      |   |   |-- karel_gym.py     # Defines KarelGymEnv
      |   |-- karel_tasks/
      |   |   |-- __init__.py
      |   |   |-- maze.py          # Defines Maze, MazeSparse, MazeWide, etc.
      |   |-- karel/
      |   |   |-- __init__.py
      |   |   |-- environment.py   # Defines KarelEnvironment and features
      |   |-- base/
      |       |-- __init__.py
      |       |-- task.py          # Defines BaseTask
      |...

Usage Summary:
  1. The FunSearch framework "evolves" a Python function `get_action(env)` to maximize `evaluate(n)`.
  2. `evaluate(n)` runs n episodes of the Karel SparseMaze environment, each seeded differently, with different locations for walls and goal.
  3. Each episode calls `run_episode()`, which repeatedly:
     - Queries `get_action(env)` to obtain actions: {0..4}.
     - Steps the Gym environment and accumulates sparse/dense rewards.
     - Terminates when Karel reaches the goal or max_steps is reached.
  4. Maze-classes in karel_tasks/maze.py carve out a random maze layout (via DFS), set a goal marker,
     and compute rewards either sparsely (1 upon reach) or densely (normalized distance progress).
  5. KarelGymEnv wraps these Tasks into a standard Gym API: it exposes `step()`, `reset()`, `render()`,
     `action_space`, `observation_space`, and handles "multiple initial configurations" if requested.
"""
\end{lstlisting}

\end{document}